\documentclass[journal]{IEEEtran}
\usepackage{amsmath,amsfonts}
\usepackage{algorithmic}
\usepackage{array}
\usepackage{textcomp}
\usepackage{stfloats}
\usepackage{url}
\usepackage{verbatim}
\usepackage{graphicx}

\usepackage{amsmath}
\usepackage{threeparttable}
\usepackage{booktabs}
\usepackage{makecell}
\usepackage{multirow}
\usepackage{bbding}
\usepackage[square,sort,comma,numbers]{natbib}

\usepackage{amssymb}
\usepackage[table]{xcolor}

\usepackage{caption}     
\usepackage{subcaption}  

\def\BibTeX{{\rm B\kern-.05em{\sc i\kern-.025em b}\kern-.08em
    T\kern-.1667em\lower.7ex\hbox{E}\kern-.125emX}}
\usepackage{balance}
\begin{document}
\title{Enhancing Audio-Visual Spiking Neural Networks through Semantic-Alignment and Cross-Modal Residual Learning}
\author{Xiang He\textsuperscript{*} \thanks{Xiang He is with the Brain-inspired Cognitive Intelligence Lab, Institute of Automation, Chinese Academy of Sciences, Beijing 100190, China, and School of Artificial Intelligence, University of Chinese Academy of Sciences, Beijing 100049, China.},
Dongcheng Zhao\textsuperscript{*} \thanks{Dongcheng Zhao is with the Brain-inspired Cognitive Intelligence Lab, Institute of Automation, Chinese Academy of Sciences, Beijing 100190, China.},
Yiting Dong \thanks{Yiting Dong and Guobin Shen are with the Brain-inspired Cognitive Intelligence Lab,Institute of Automation, Chinese Academy of Sciences, Beijing 100190, China, and School of Future Technology, University of Chinese Academy of Sciences, Beijing 100049, China.},
Guobin Shen, 
Xin Yang, \thanks{Xin Yang is with the CAS Key Laboratory of Molecular Imaging, Institute of Automation, Chinese Academy of Sciences, Beijing 100190, China.}
Yi Zeng \thanks{Yi Zeng is with the Brain-inspired Cognitive Intelligence Lab, Institute of Automation, Chinese Academy of Sciences, Beijing 100190, China, and Center for Long-term Artificial Intelligence, Beijing 100190, China, and University of Chinese Academy of Sciences, Beijing 100049, China, and Key Laboratory of Brain Cognition and Brain-inspired Intelligence Technology, Chinese Academy of Sciences, Shanghai, 200031, China. }

\thanks{
  \textsuperscript{*}These authors contributed equally.}

\thanks{
  The corresponding author is Xin Yang (e-mail: xin.yang@ia.ac.cn) and Yi Zeng (e-mail: yi.zeng@ia.ac.cn).
}

}

\markboth{Journal of \LaTeX\ Class Files,~Vol.~18, No.~9, September~2020}%
{How to Use the IEEEtran \LaTeX \ Templates}

\maketitle

\begin{abstract}
    Humans interpret and perceive the world by integrating sensory information from multiple modalities, such as vision and hearing. Spiking Neural Networks (SNNs), as brain-inspired computational models, exhibit unique advantages in emulating the brain’s information processing mechanisms. However, existing SNN models primarily focus on unimodal processing and lack efficient cross-modal information fusion, thereby limiting their effectiveness in real-world multimodal scenarios.
    To address this challenge, we propose a semantic-alignment cross-modal residual learning (S-CMRL) framework, a Transformer-based multimodal SNN architecture designed for effective audio-visual integration. S-CMRL leverages a spatiotemporal spiking attention mechanism to extract complementary features across modalities, and incorporates a cross-modal residual learning strategy to enhance feature integration. Additionally, a semantic alignment optimization mechanism is introduced to align cross-modal features within a shared semantic space, improving their consistency and complementarity. Extensive experiments on three benchmark datasets CREMA-D, UrbanSound8K-AV, and MNISTDVS-NTIDIGITS demonstrate that S-CMRL significantly outperforms existing multimodal SNN methods, achieving the state-of-the-art performance. The code is publicly available at \texttt{https://github.com/Brain-Cog-Lab/S-CMRL}.
\end{abstract}


\section{INTRODUCTION}

Human perception of the external world arises from the integration of information across multiple modalities, including vision, hearing, and language. Compared to unimodal perception, multimodal learning provides a richer and more comprehensive representation of information. Furthermore, the integration of multiple modalities enhances perceptual robustness, facilitating a deeper understanding of the environment~\cite{ernst2004merging, noppeney2021perceptual}. 
Among these modalities, vision and hearing serve as the two primary sensory pathways for acquiring external information~\cite{enoch2019evaluating, bulkin2006seeing}, and their integration plays a pivotal role in daily life. For example, in low-light environments, auditory cues can compensate for insufficient visual information, enabling a more accurate perception of the external environment and reducing uncertainty. In recent years, audio-visual multimodal learning has achieved remarkable advancements in various applications, such as audio-visual speech recognition~\cite{kim2021cromm, peng2022balanced, yeo2024akvsr}, video sound separation~\cite{gan2020music}, video sound source localization~\cite{hu2020discriminative}, and audio-visual event localization~\cite{lin2019dual, feng2023css, he2024cace}. These developments highlight the importance of leveraging complementary information from both visual and auditory modalities to address complex challenges in perception.

\begin{figure}[t]
	\centering
		\includegraphics[width=1.0\linewidth]{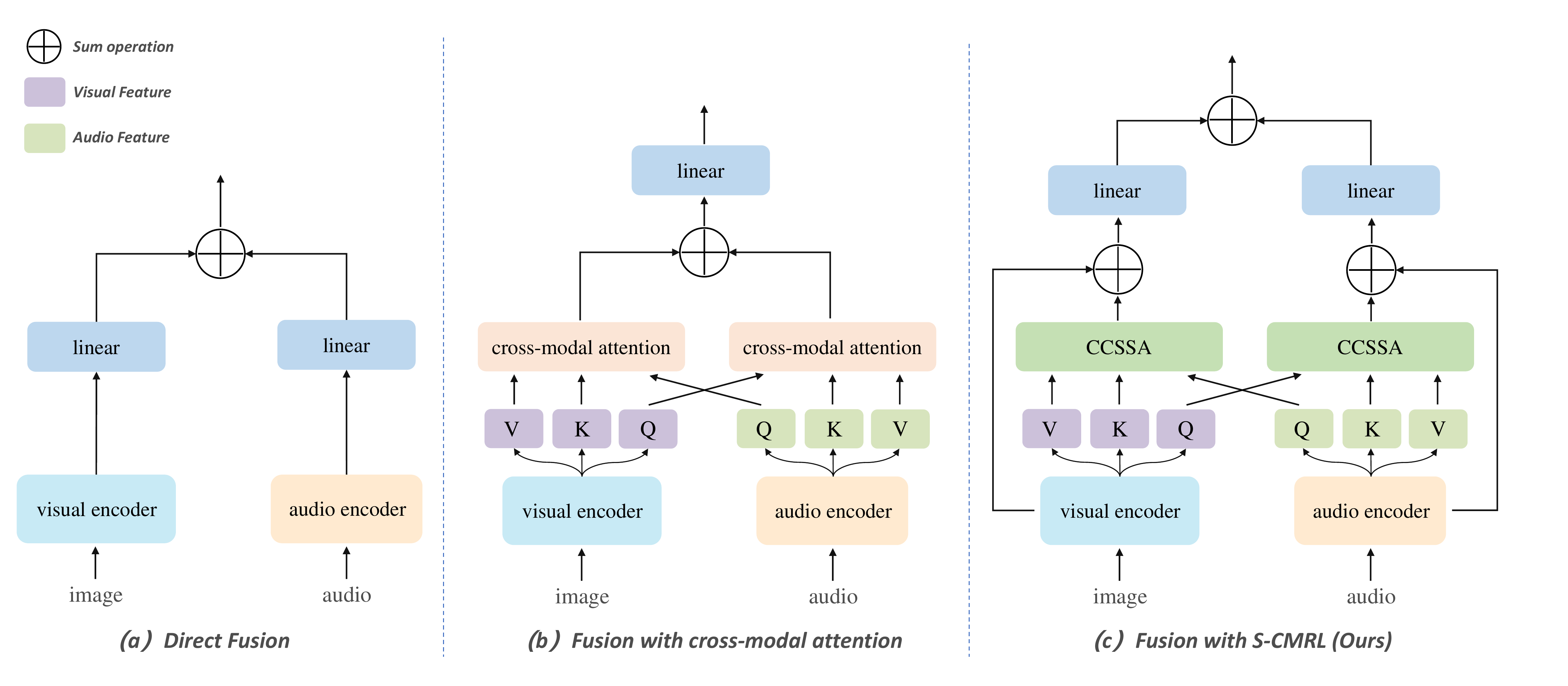}
	\caption{ 
  Different cross-modal fusion methods in spiking neural networks. (a) Direct fusion, which typically sums the features from different modalities directly. (b) Fusion with cross-modal attention mechanisms. (c) Our proposed semantic-alignment cross-modal residual learning fusion.
  ``Q'': Query embedding; ``K'': Key embedding; ``V'': Value embedding.}
	\label{fig_compare}
\end{figure}

Spiking Neural Networks (SNNs), inspired by biological nervous systems, provide a compelling computational paradigm characterized by event-driven information processing and sparse activation. Unlike traditional Artificial Neural Networks (ANNs), SNNs transmit information through discrete spike sequences. The sparse spikes and event-driven computation paradigm inherent to SNNs significantly reduces power consumption. Similar to the brain, these spikes encode information temporally, enabling SNNs to handle spatiotemporal data.
In recent years, SNNs have demonstrated impressive performance across diverse applications, including computer vision~\cite{deng2022temporal, zhou2022spikformer, li2022spike, shen2023brain, feng2024spiking, xie2024eisnet}, natural language processing~\cite{xiao2022towards, shen2023astrocyte, su2024snn}, and audio processing~\cite{pan2021multi, wang2024spikevoice, yang2024svad}.

Despite significant progress in SNN research, most existing models focus on unimodal processing, with limited exploration of multimodal SNNs. 
In contrast to human cognition, where multimodal integration is fundamental~\cite{zaadnoordijk2022lessons, lin2023multimodality}, 
existing multimodal SNN approaches often fall short in effectively integrating multimodal information. Some studies simply combine the features of two modalities in a straightforward manner~\cite{zhang2020efficient, liu2022event}, while others overlook the distinct characteristics of auditory and visual modalities and lack sufficient exploration of the complementary information between modalities~\cite{guo2023transformer, jiang2023cmci}, as illustrated in Fig.~\ref{fig_compare}(a) and Fig.~\ref{fig_compare}(b). These methods do not fully exploit intermodal complementarity, which limits their effectiveness in multimodal learning.

\begin{figure}[t]
	\centering
		\includegraphics[width=0.9\linewidth]{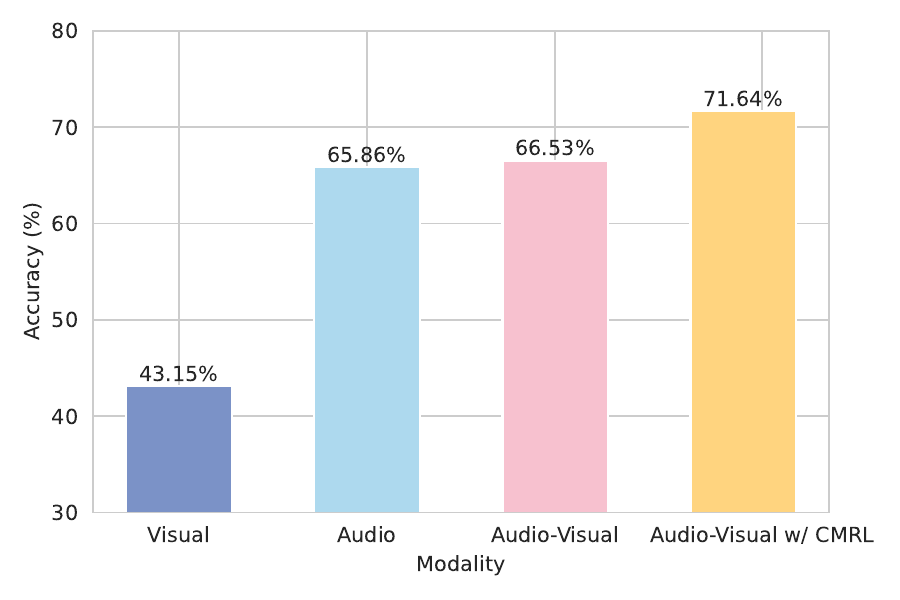}
	\caption{ 
  Experimental results of the Spiking Transformer on the CRMEA-D dataset.
  CMRL represents Cross-Modal Residual Learning. Due to the weaker visual signals compared to audio in CREMA-D, traditional cross-modal fusion strategies show limited improvement. When incorporating cross-modal features as residuals into the unimodal representations, model performance improves. This highlights the importance of preserving the unimodal-specific semantic features.}
	\label{with_res}
\end{figure}

A fundamental challenge in multimodal learning arises from the inherent differences in feature distributions across modalities. Consequently, a naive direct fusion of these features often results in conflicts between modalities. For example, when the visual data is less informative in some situations while auditory data is more discriminative, the introduced cross-modal features may be interfered by low-quality visual information, leading to suboptimal performance. To verify this, we select the cross-modal attention method proposed by Guo et al.~\cite{guo2023transformer} to extract cross-modal features. Subsequently, we regard the cross-modal features as “complementary” to the unimodal features and fuse them as residuals with the original unimodal features. We conduct experiments using the spiking Transformer in the CRMEA-D~\cite{cao2014crema} dataset, as shown in Fig.~\ref{with_res}. The experimental results show that preserving the unimodal features while incorporating cross-modal residuals can achieve better model performance. This result highlights the importance of preserving modality-specific representations while leveraging complementary intermodal information to achieve efficient audio-visual modal integration.

To address the above challenge, we propose a \textbf{S}emantic-Alignment \textbf{C}ross-\textbf{M}odal \textbf{R}esidual \textbf{L}earning (S-CMRL) framework, which is based on a Transformer-driven multimodal spiking neural network for audio-visual learning. As illustrated in Fig.~\ref{fig_compare}(c), the proposed framework encodes audio and visual data as sequential inputs and employs a cross-modal spiking attention mechanisms to introduce semantic information that guides the residual learning of cross-modal features.
By effectively integrating cross-modal complementary information, S-CMRL enhances the intermodal collaboration while mitigating conflicts that arise from direct fusion. 

Specifically, our framework consists of two primary modules. 
First, we propose a cross-modal complementary spatiotemporal spiking attention (CCSSA) module. This module extends the traditional unimodal spiking Transformer to accommodate dual-modal inputs and extract complementary semantic information from another modality. These complementary features, treated as “residuals,” are fused into the original modality’s features, allowing effective cross-modal integration while preserving modality-specific characteristics. This approach mitigates conflicts from simple fusion methods and enhances the network’s ability to address complex scenarios. Second, we introduce a semantic alignment optimization (SAO) mechanism to refine cross-modal residual features. By aligning cross-modal features from the same category across visual and auditory modalities within a shared semantic space, this mechanism reinforces consistency and improves the quality of complementary representations. 

We evaluate our framework on three audio-visual datasets CREMA-D and UrbanSound8K-AV, as well as the neuromorphic dataset MNISTDVS-NTIDIGITS. Experimental results demonstrate that our multimodal SNN achieves state-of-the-art performance on all three datasets, while exhibiting strong robustness under noisy conditions. 

In summary, the main contributions of this paper can be summarized as follows:

\begin{itemize} 
  \item We propose a novel cross-modal complementary spatiotemporal spiking attention mechanism, which effectively integrates cross-modal complementary information while preserving modality-specific semantic information, thereby enhancing representational expressiveness.
  \item We propose a semantic alignment optimization mechanism to align cross-modal features within a shared semantic space, improving cross-modal feature consistency and overall multimodal learning performance. 
  \item Based on these two modules, we construct a semantic-alignment cross-modal residual learning framework for multimodal SNNs. This framework provides an efficient feature fusion strategy and achieves state-of-the-art performance on three public datasets, demonstrating superior accuracy and robustness compared to existing methods. 
\end{itemize}

\section{RELATED WORK}
\subsection{Audio-Visual Learning}
In neural networks, multimodal fusion integrates two or more modalities to tackle complex tasks and improve model accuracy and generalization. Audio-visual learning primarily focuses on uncovering relationships between visual and auditory modalities, with feature fusion as the central research focus. Hu et al.~\cite{hu2016temporal} and Yang et al.~\cite{yang2020telling} explored feature concatenation, fusing audio-visual features along specific dimensions to generate a unified feature vector. Moving beyond simple concatenation, Wu et al.~\cite{wu2019dual} proposed a dual attention matching module to facilitate higher-level event information modeling. Moreover, certain studies addressed the issue of single-modal imbalance~\cite{feichtenhofer2019slowfast, peng2022balanced} by strengthening cross-modal alignment and reducing inter-modal discrepancies to optimize fusion. Yeo et al.~\cite{yeo2024akvsr} tackled the problem of insufficient speech information in the visual modality by leveraging the audio modality as a complementary source. In audio-visual modalities, the visual modality typically provides spatial information, whereas the auditory component captures temporal dynamics; each modality thus presents distinct representational patterns and semantic information. Consequently, designing a network that can dynamically fuse these complementary elements has become a pivotal challenge in audio-visual learning research.

\subsection{Unimodal Spiking Neural Networks}


Most existing research on spiking neural networks (SNNs) focuses on single-modal tasks. 
For vision-related tasks, Wu et al.~\cite{wu2018spatio} introduced gradient approximation of spike functions for gradient computation, applying backpropagation in both the spatial and temporal domains, namely Spatio-Temporal Backpropagation (STBP). In subsequent work~\cite{wu2019direct}, they presented a discrete and iterative form of the commonly used LIF neuron model~\cite{dayan2005theoretical}. Deng et al.~\cite{deng2022temporal} proposed a temporal-efficient training approach that converges SNNs to flatter minima. Zhou et al.~\cite{zhou2022spikformer} introduced Spikformer, incorporating a spiking self-attention mechanism that leverages spike-based queries, keys, and values to capture sparse visual features.
In audio-related tasks, Auge et al.~\cite{auge2021end} employed resonator neurons as the input layer of an SNN for online audio classification. Yu et al.~\cite{yu2022stsc} proposed a spatiotemporal synaptic connection module composed of a temporal response filter module and a feedforward lateral inhibition module, demonstrating its efficacy in spoken digit recognition.
Although unimodal approaches have achieved significant progress, the rising importance of multimodal learning highlights the need to integrate diverse modalities to enhance the representational capacity and generalization of models. Consequently, extending traditional unimodal SNN techniques to multimodal contexts is emerging as a key challenge in the current and future development of spiking neural networks.

\subsection{Audio-Visual Spiking Neural Networks}
Research on audio visual spiking neural networks (AV-SNNs) remains limited. Early studies primarily focused on simple connections or straightforward combinations of visual and auditory modality features. Zhang et al.~\cite{zhang2020efficient} employed excitatory and inhibitory lateral connections to facilitate cross-modal coupling in SNNs trained on individual modalities. Liu et al.~\cite{liu2022event} introduced an attention-based cross-modal network that leverages an attention mechanism to weigh each modality's contribution, enabling cross-modal fusion.
More recently, researchers have developed advanced multimodal fusion methods to improve AV-SNNs. Guo et al.~\cite{guo2023transformer} integrated SNNs with Transformers, combining unimodal sub-networks for visual and auditory modalities and proposing a novel spiking cross-attention module for audio-visual classification. Jiang et al.~\cite{jiang2023cmci} proposed a cross-modal current integration module that fuses SNNs from different modalities at either the feature or decision level.
However, existing approaches overlook the unique characteristics of auditory and visual modalities and their complementary interactions. In this work, we preserve the distinct features of each modality and design mechanisms that enable each unimodal branch to leverage complementary information from the other. This approach achieves more effective cross-modal fusion and enhances multimodal learning performance.

\section{Preliminary}
In this section, we first formally define the research problem and introduce the key neuron model used in this study, namely the ``Leaky Integrate-and-Fire (LIF) Neuron'', and the crucial network model, the ``Spiking Transformer''. These components are fundamental to efficient multimodal data fusion.

\subsection{Problem Definition}
For a given multimodal dataset $\mathcal{D}$, it can be expressed as $\mathcal{D}=\left\{\left(\boldsymbol{x}_{i}^{a}, \boldsymbol{x}_{i}^{v}, y_{i}\right)\right\}_{i=1}^{n_ t}$, where $\boldsymbol{x}_{i}^{a} \in \mathcal{X}^a$ and $\boldsymbol{x}_{i}^{v} \in \mathcal{X}^v$ denote the input data of the audio modality and visual modality, respectively. $y_{i} \in \mathcal{Y}$ is the corresponding label and $n_t$ is the total number of training samples for the corresponding task.
The objective of our research is to learn a multimodal model with the parameter $\theta$, 
denoted as $f_{\boldsymbol{\theta}}$, to predict class labels from audio-visual inputs:
\begin{equation}
    f_{\boldsymbol{\theta}} \colon \mathcal{X}^a \times \mathcal{X}^v \rightarrow \mathcal{Y}.
\end{equation}

The model is optimized by minimizing the expected risk based on the cross-entropy loss function $\mathcal{L}_{ce}$:
\begin{equation}
    \underset{\boldsymbol{\theta}}{\operatorname{argmin}} \mathbb{E}_{\left(\boldsymbol{x}^a, \boldsymbol{x}^v, y\right) \sim \mathcal{D}} \left[\mathcal{L}_{ce}\left(f_{\boldsymbol{\theta}}\left(\boldsymbol{x}^a, \boldsymbol{x}^v\right), y\right)\right].
\end{equation}
In this study, our primary focus is on designing the model $f_{\boldsymbol{\theta}}$ to achieve efficient cross-modal feature fusion. To this end, we leverage a combination of SNNs and Transformer architectures for the model $f_{\boldsymbol{\theta}}$, aiming to enhance the spatiotemporal modeling capabilities of multimodal feature fusion through efficient spike-based information processing.

\subsection{Leaky Integrate-and-Fire (LIF) Neuron}
The Leaky Integrate-and-Fire (LIF) neuron is a fundamental component of SNNs. Its membrane potential increases with the accumulation of the input current and leaks gradually over time. When the potential reaches a certain threshold, the neuron emits a spike, and subsequently the membrane potential resets to the resting potential $V_{reset}$. With the resting potential $V_{reset}$ set to $0$, the membrane potential update equation of the LIF model can be expressed in the following discrete form:
\begin{gather}
    \boldsymbol{V}^{l}(t) = \boldsymbol{V}^{l} (t-1) + \frac{1}{\tau}
    \left(\boldsymbol{W}^l \boldsymbol{S}^{l-1}(t) - \boldsymbol{V}^{l} (t-1)\right), \\
    \boldsymbol{S}^{l}(t) = \Theta\left(\boldsymbol{V}^{l}(t) - V_{th}\right),\\
    \boldsymbol{V}^{l}(t) = \boldsymbol{V}^{l}(t) \cdot \left(1 - \boldsymbol{S}^{l}(t)\right),
    \label{eq1}
\end{gather}
where $\tau$ is the leakage factor and $\boldsymbol{V}^{l}(t)$ denotes the membrane potential of the neuron in layer $l$ at time step $t$. $\boldsymbol{W}^l$ and $\boldsymbol{S}^l$ denote the weight matrix of layer $l$ and the spikes fired in layer $l$, respectively. The $\Theta$ is the Heaviside step function. In our study, the leakage factor $\tau$ is set to 2.0 and the threshold $V_{th}$ is set to 1.0.

\subsection{Spiking Transformer}
SNNs excel in temporal information modeling and energy-efficient computation, while the Transformer architecture is well known for its ability to capture long-range dependencies. To harness the advantages of both, we choose the Spiking Transformer as our backbone model. Specifically, we integrate the Spiking Patch Splitting (SPS) and Spiking Self-Attention (SSA) mechanisms, as proposed in~\cite{zhou2022spikformer}, to improve multimodal feature representation.

In SPS module, input data from both the audio and visual modalities are encoded and projected into a $D$-dimensional spiking feature space, where they are partitioned into fixed-size feature patches of size $N$. Each SPS module consists of multiple submodules, each comprising a convolutional layer, batch normalization, LIF neurons, and a max-pooling layer. After passing the SPS module, the audio and visual inputs, $\boldsymbol{x}^a$ and $\boldsymbol{x}^v$, are transformed into patch sequences:
\begin{equation}
  \boldsymbol{x}^a = \mathcal{SPS}_a (\boldsymbol{x}^a), \quad \boldsymbol{x}^v = \mathcal{SPS}_v (\boldsymbol{x}^v),
\end{equation}
where $\boldsymbol{x}^a, \boldsymbol{x}^v \in \mathbb{R}^{B \times T \times N \times D}$, and $B$, $T$, $N$, and $D$ denote the batch size, temporal length, spatial locations, and feature dimensions, respectively..

Spiking Self-Attention (SSA) is a spike-based variant of the conventional self-attention mechanism, designed to bridge the incompatibility between standard attention operations and spiking attention. In spiking self-attention, queries, keys, and values are represented in a pure spike format. Following the formulation in~\cite{zhou2022spikformer}, the computation is performed as follows:
\begin{equation}
  \begin{gathered}
      Q = \mathcal{S} \mathcal{N}^Q\left(\operatorname{BN}\left(\boldsymbol{x} W^Q\right)\right),\\
      K = \mathcal{S N}^K\left(\operatorname{BN}\left(\boldsymbol{x} W^K\right)\right), \\
      V = \mathcal{S} \mathcal{N}^V\left(\operatorname{BN}\left(\boldsymbol{x} W^V\right)\right), \\
      \operatorname{SSA}(\boldsymbol{x}) = \mathcal{S N}\left(\operatorname{BN}\left(\operatorname{Linear}\left(\mathcal{S N}\left(Q K^{\mathrm{T}} V \cdot s\right) \right)\right)\right),
  \end{gathered}
  \label{ssa}
  \end{equation}
where $Q$, $K$, and $V$ denote the query, key, and value, respectively; $s$ is the scaling factor; $\operatorname{Linear}$ represents the linear transformation layer; BN denotes Batch Normalization; and $\mathcal{SN}$ stands for the spike neuron layer. Notably, spiking self-attention does not change the dimension of the input features. Thus, the output satisfies $\operatorname{SSA}(\boldsymbol{x}) \in \mathbb{R}^{B \times T \times N \times D}$.

By combining SPS and SSA, the Spiking Transformer effectively leverages the sparse activation characteristic of spike neurons while preserving the powerful spatiotemporal information modeling capabilities inherent to the Transformer architecture. This fusion provides efficient and precise representations for multimodal learning.

\section{Methods}
A key challenge in multimodal learning is how to efficiently integrate cross-modal information to fully leverage its complementary characteristics. Traditional approaches typically employ feature concatenation or simple aggregation for classification. For naive multimodal fusion, where different modality features are directly summed, the Spiking Transformer model $f_{\boldsymbol{\theta}}$ can be formulated as:
\begin{gather}
    \boldsymbol{z}^a = g_a(\boldsymbol{x}^a)= \operatorname{SSA}_a(\boldsymbol{x}^a), \;
    \boldsymbol{z}^v = g_v(\boldsymbol{x}^v) = \operatorname{SSA}_v(\boldsymbol{x}^v), \label{eq:12}\\
    f_{\boldsymbol{\theta}}\left(\boldsymbol{x}^a, \boldsymbol{x}^v\right) = h \circ \Gamma\left(\operatorname{MLP}_a(\boldsymbol{z}^a) + \operatorname{MLP}_v(\boldsymbol{z}^v)\right),
\end{gather}
where $g: \mathcal{X} \rightarrow \mathcal{Z}$ represents the mapping from the input space $\mathcal{X}$ to the feature space $\mathcal{Z}$, $\operatorname{MLP}$ refers to a multilayer perceptron layer, and $\Gamma$ denotes global average pooling. The function $h: \mathcal{Z} \rightarrow \mathcal{Y}$ maps features to output predictions, typically implemented as a linear layer.

However, this fusion strategy overlooks the complex spatiotemporal dependencies between modalities, limiting the model’s representation capacity. Moreover, the lack of explicit constraints on cross-modal features can result in semantic mismatch, reducing the effectiveness of fusion. 

To overcome these challenges, we propose the semantic-alignment cross-modal residual learning framework, which consists of two primary modules. These two modules are described in detail in the following section.

\begin{figure}[t]
	\centering
		\includegraphics[width=1.0\linewidth]{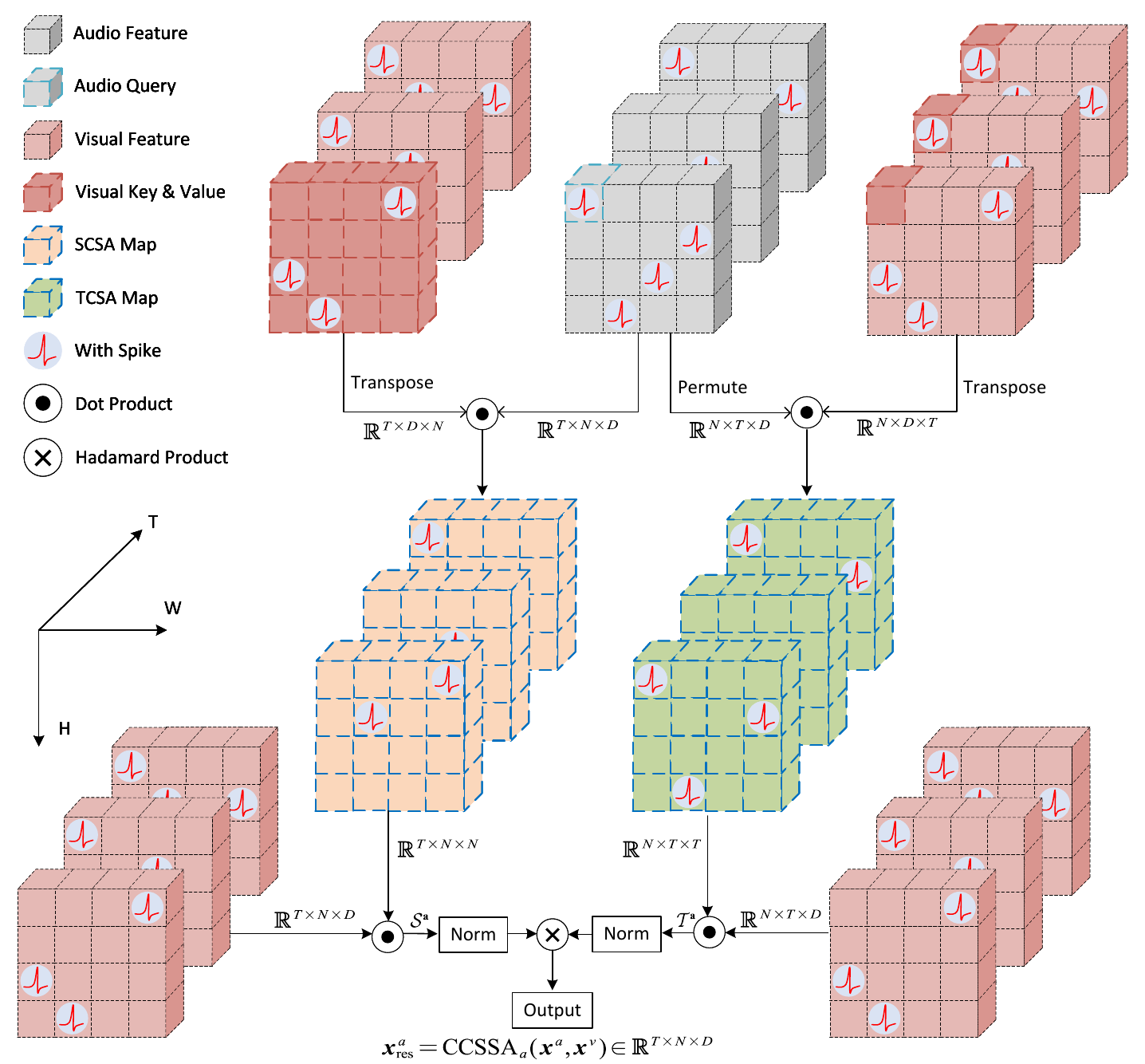}
	\caption{ 
  Schematic of cross-modal complementary spatio-temporal spiking attention, using the computation process of the complementary feature $\boldsymbol{x}^a_{\text{res}}$ in audio features as an example. Best viewed in color.
  }
	\label{figccssa}
\end{figure}

\subsection{Cross-modal Complementary Spatiotemporal Spiking Attention}
A major limitation of existing multimodal fusion approaches is that their feature mapping functions $g$ process each modality independently, as illustrated in Eq.~\ref{eq:12}, failing to explicitly model cross-modal complementary relationships. To overcome this constraint, we propose the cross-modal complementary spatiotemporal spiking attention (CCSSA) mechanism, as shown in Fig.~\ref{figccssa}, which enables each modality to dynamically integrate complementary information from the other modality and improves the representational capacity of the original modality.
Specifically, the complementary features from the other modality are treated as residuals and are fused with the distinctive features of the original modality. This process can be mathematically formulated as:
\begin{gather}
    g_a(\boldsymbol{x}^a)= \boldsymbol{x}^a + \alpha \cdot \operatorname{CCSSA}_a(\boldsymbol{x}^a, \boldsymbol{x}^v) ,\label{eq14}\\
    g_v(\boldsymbol{x}^v)= \boldsymbol{x}^v + \alpha \cdot \operatorname{CCSSA}_v(\boldsymbol{x}^v, \boldsymbol{x}^a) ,\label{eq15}
\end{gather}
where $\alpha$ is a hyperparameter to control the fusion strength of complementary information.

In our approach, CCSSA not only captures complementary information in the spatial dimension but also models complementary relationships in the temporal dimension, thereby enables effective cross-modal spatio-temporal information fusion. CCSSA is composed of two key components: spatial complementary spiking attention (SCSA) and temporal complementary spiking attention (TCSA). Let the auditory and visual input features be denoted as $\boldsymbol{x}^a$ and $\boldsymbol{x}^v$ respectively, each of dimension $\mathbb{R}^{T \times B \times N \times D}$. In the following, we will take the computation process of $\operatorname{CCSSA}_a(\boldsymbol{x}^a, \boldsymbol{x}^v)$ as an example to describe the computation of SCSA and TCSA in detail.

\paragraph{Spatial Complementary Spiking Attention (SCSA)}
SCSA is designed to capture complementary information in the spatial dimension. It enhances information fusion across modalities by modeling the correlations between spatial locations. First, we compute the complementary features $\mathcal{S}^a$ of cross-modal spatial attention according to the Eq.~\ref{ssa}, where the input to the query $Q_s$ is the audio modality feature $\boldsymbol{x}^a$ and the inputs to the key $K_s$ and value $V_s$ are the visual modality features $\boldsymbol{x}^v$. The $\mathcal{S}^a$ represents the mapping of the query audio feature representation in the visual modality to capture the interrelated information between audio and vision.

Subsequently, the output of spatial attention is spatially normalized to obtain a compact spatial feature representation:
\begin{equation}
    \mathbf{\mathcal{S}^a}_{\text{reduced}} = \frac{1}{N} \sum_{i=1}^{N} \mathbf{\mathcal{S}^a}_{:, :, i, :} \; \in \mathbb{R}^{T \times B \times D},
\end{equation}
where $N$ denotes the size of the spatial dimension, and $\mathbf{\mathcal{S}^a}_{:, :, i, :}$ denotes the feature of the $i$-th spatial location. Through the averaging operation, $\mathbf{\mathcal{S}^a}_{\text{reduced}}$ aggregates the information from the spatial dimension to the temporal and batch dimensions to form a compact feature representation.

To ensure information integrity and compatibility with subsequent processing, we expand the normalized features back to the original spatial dimension:
\begin{equation}
    \mathbf{\mathcal{S}^a}_{\text{expanded}} = \mathbf{\mathcal{S}^a}_{\text{reduced}} \otimes \mathbf{1}_N \; \in \mathbb{R}^{T \times B \times N \times D},
\end{equation}
where $\mathbf{1}_N$ denotes an all-ones vector of size $N$, and $\otimes$ denotes the outer product operation. This expansion ensures that $\mathbf{\mathcal{S}^a}_{\text{expanded}}$ retains the compactness of the normalized features while restoring the original spatial resolution for subsequent cross-modal fusion.

\begin{figure*}[t]
  \centering
    \includegraphics[width=0.95\linewidth]{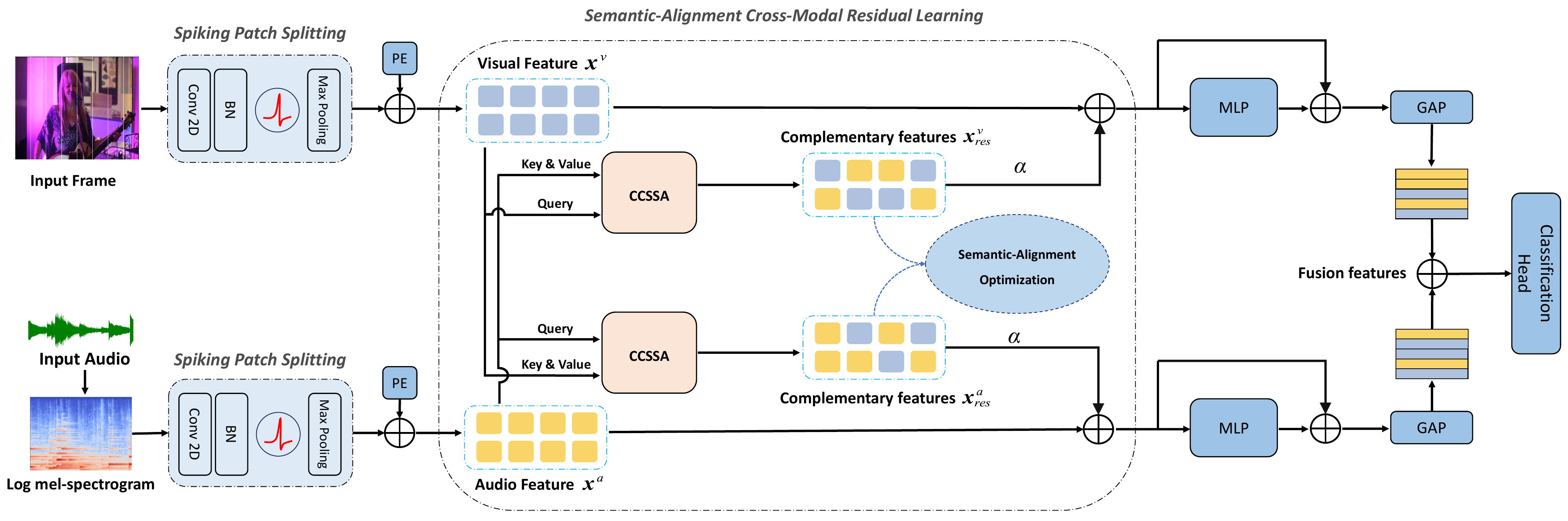}
  \caption{ 
  Overview of proposed semantic-alignment cross-modal residual learning framework. The network processes visual and auditory inputs through independent pathways. Following positional embedding, These pathways converge in a central module, which employs a novel cross-modal complementary spatiotemporal spike attention mechanism. This mechanism effectively exploits complementary information between modalities and integrates it as residuals into the unique feature representations of each modality. Additionally, the semantic alignment optimization further enhances the consistency of cross-modal features.}
  \label{fig}
\end{figure*}

\paragraph{Temporal Complementary Spiking Attention (TCSA)}
TCSA is designed to extract complementary information along the temporal dimension. By modeling temporal correlations, it facilitates the integration of temporal information across different modalities and enhances the model’s capacity to capture temporal dynamics. During computation, the input features are first rearranged to accommodate temporal attention calculations as $\boldsymbol{x}^a \in \mathbb{R}^{B \times N \times T \times D}$ and $\boldsymbol{x}^v \in \mathbb{R}^{B \times N \times T \times D}$. 
We then compute the complementary features $\mathcal{T}^a$ of cross-modal temporal attention using Eq.~\ref{ssa}.
After obtaining the temporal complementary feature $\mathbf{\mathcal{T}^a}$, we perform a dimensional transformation on it so that it becomes $\mathbf{\mathcal{T}^a} \in \mathbb{R}^{T \times B \times N \times D}$.

Subsequently, the temporal dimension is normalized to obtain a compact representation with temporal features at each batch and spatial location:
\begin{equation}
    \mathbf{\mathcal{T}^a}_{\text{reduced}} = \frac{1}{T} \sum_{j=1}^{T} \mathbf{\mathcal{T}^a}_{j, :, :, :} \; \in \mathbb{R}^{B \times N \times D},
\end{equation}
where $T$ is the time dimension length and $\mathbf{\mathcal{T}^a}_{j, :, :, :}$ represents the features at the $j$-th time step. 
Next, the normalized features are expanded back to the original time dimension for fusion with subsequent processing modules:
\begin{equation}
    \mathbf{\mathcal{T}^a}_{\text{expanded}} = \mathbf{\mathcal{T}^a}_{\text{reduced}} \otimes \mathbf{1}_T \; \in \mathbb{R}^{T \times B \times N \times D}.
\end{equation}
where $\mathbf{1}_T$ denotes an all-ones vector of size $T$.

\paragraph{Spatiotemporal Complementary Fusion}
With complementary information in spatial and temporal dimensions, we fuse them through element-wise multiplication:
\begin{equation}
    \operatorname{CCSSA}_a(\boldsymbol{x}^a, \boldsymbol{x}^v) = \mathbf{\mathcal{S}^a}_{\text{expanded}} \ast \mathbf{\mathcal{T}^a}_{\text{expanded}} \; \in \mathbb{R}^{T \times B \times N \times D}.
\end{equation}
Finally, the fused complementary feature $\operatorname{CCSSA}_a(\boldsymbol{x}^a, \boldsymbol{x}^v)$ is incorporated as residual into the original modality-specific feature, enhancing the expressiveness of cross-modal information.

\subsection{Semantic Alignment Optimization}
Within the CCSSA mechanism, we obtain cross-modal complementary features $\boldsymbol{x}^a_{\text{res}} = \operatorname{CCSSA}_a(\boldsymbol{x}^a, \boldsymbol{x}^v)$ and $\boldsymbol{x}^v_{\text{res}} = \operatorname{CCSSA}_v(\boldsymbol{x}^v, \boldsymbol{x}^a)$. Although CCSSA enables complementary feature fusion across spatial and temporal dimensions, cross-modal semantic shift may still persists. Specifically, the primary cause of cross-modal semantic shift lies in the inherent distribution differences between modalities. Due to the unique feature representations of each modality, their features often exhibit inconsistencies in distribution within the shared semantic space. This distributional inconsistency makes it difficult for the model to learn a stable cross-modal semantic mapping in a unified feature space, thus impairing the effectiveness of multimodal fusion.

To address this issue, we propose semantic alignment optimization (SAO), which explicitly aligns cross-modal features within a shared semantic space. SAO aims to improve semantic consistency and strengthen multimodal representations, which is implemented through the following loss function:
\begin{equation}
    \begin{gathered}
    \mathcal{L}_{\text {sao}} =\frac{1}{B} \sum_{i =1}^B \frac{1}{T} \sum_{t =1}^T-\log \left\{\frac{\exp \left(\boldsymbol{x}^{a,t}_{\text{res}, i} \cdot \boldsymbol{x}^{v,t}_{\text{res}, i} / \tau\right)}{\sum_{j=1}^B \exp \left(\boldsymbol{x}^{a,t}_{\text{res}, i} \cdot \boldsymbol{x}^{v,t}_{\text{res}, i} / \tau\right)}\right\},
    \end{gathered}
  \end{equation}
  where $i \in \left\{1, 2, \cdots, B\right\}$ denotes the sample index within a batch, and $\boldsymbol{x}^{a,t}_{\text{res},i}$ represents the cross-modal complementary feature $\boldsymbol{x}^a_{\text{res}}$ of the $i$-th sample at time step $t$ in batch $B$. The parameter $\tau \in \mathbb{R}^{+}$ is a temperature coefficient used to adjust the smoothness of the distribution.
  
  The loss function $\mathcal{L}_{\text{sao}}$ aims to optimize the model's performance by explicitly aligning complementary cross-modal features at the same time step within the shared semantic space. By incorporating the SAO mechanism, the model can improve the semantic consistency of cross-modal complementary features during the multimodal fusion, leading to more robust and discriminative cross-modal representations. The final optimization objective is given by:
  \begin{equation}
      \mathcal{L} = \mathcal{L}_{ce} + \mathcal{L}_{\text {sao}}.
  \end{equation}
  
\subsection{Overall Architecture}
In this study, we propose an audiovisual multimodal spiking neural network framework, as illustrated in Fig.~\ref{fig}. This framework is designed to fully exploit the complementary information between visual and auditory modalities, thereby enhancing the robustness and discriminative ability of multimodal feature fusion. The model consists of three core modules: spiking patch splitting, cross-modal complementary spatiotemporal spiking attention, and semantic alignment optimization.


First, the raw visual and auditory inputs are processed by the SPS module, which converts sequential image frames and log-Mel spectrograms into discrete spike representations. This transformation is crucial for adapting the input data to the SNNs, enabling energy-efficient computation. The position embedded spike representations are then fed into the CCSSA module, which captures and integrates complementary information between modalities across both spatial and temporal dimensions, ensuring effective multimodal feature fusion. 

Following this, the SAO mechanism further refines and aligns cross-modal features within a shared semantic space. By explicitly aligning cross-modal features, SAO mitigates modality-specific feature interference and enhances cross-modal feature consistency. 

Finally, the fused features are passed through a classification head, which performs the final multimodal recognition task. This integrated framework leverages the complementary strengths of spiking neurons and cross-modal attention mechanisms, offering an efficient and biologically plausible solution for robust audiovisual processing.

\section{EXPERIMENTS}
In this section, we first describe the three datasets employed in our experiments and detail their corresponding experimental setups. We then conduct experiments on these datasets to compare our method with the current state-of-the-art approaches. The results demonstrate that our method outperforms existing methods under both clean and noisy conditions. Finally, we perform comprehensive ablation studies to showcase the effectiveness of each component in our proposed method.
\subsection{Datasets}
\subsubsection{CREMA-D}
CREMA-D~\cite{cao2014crema} is an audiovisual dataset for speech emotion recognition containing 7442 video clips of 2 to 3 seconds duration from 91 actors. The dataset covers the six most common emotion categories: anger, happiness, sadness, neutrality, disgust and fear. The division of the dataset follows the method of Peng et al.~\cite{peng2022balanced}, which randomly divides the dataset into training and validation sets, as well as a test set, in a ratio of 9:1. Ultimately, the training and validation sets contain 6698 samples, and the test set contains 744 samples.

\subsubsection{UrbanSound8K-AV}
The UrbanSound8K-AV dataset ~\cite{guo2023transformer} is a combination of the UrbanSound8K audio dataset ~\cite{salamon2014dataset} and its corresponding image dataset. The UrbanSound8K-AV dataset contains the same number of samples as the UrbanSound8K audio dataset, totaling 8732 audiovisual samples. Each sample consists of a high-resolution color image and a 4-second audio signal. We follow Guo et al.~\cite{guo2023transformer} by randomly dividing the dataset into training and test sets in a 7:3 ratio.

\subsubsection{MNISTDVS-NTIDIGITS}
MNISTDVS-NTIDIGITS is a audio-visual dataset spliced together from the MNISTDVS dataset \cite{serrano2013128} and the NTIDIGITS dataset \cite{serrano2013128}. Unlike traditional sensors, these datasets are spatio-temporal event data collected by dynamic visual sensors (DVS) and dynamic audio sensors (DAS), also known as neuromorphic datasets. Following Liu et al.~\cite{liu2022event}, we select 10 numerical categories in N-TIDIGITS (“zero”, “1” to “9 “), a total of 4500 audio samples are reused to match the MNIST-DVS dataset, resulting in 10000 audiovisual samples. The dataset is divided into training and test sets in a 5:5 ratio.
\begin{table*}[t]
  \centering
  \begin{threeparttable}
    \resizebox{1.0\linewidth}{!}{
      \begin{tabular}{cccccc}
        \toprule
        \textbf{Dataset} &\textbf{Category} & \textbf{Methods} & \textbf{Architecture} & \textbf{T} & \textbf{Accuracy}\\
        \midrule
        \multirow{7}{*}{\text { CRMEA-D }} &\multirow{4}{*}{\text { ANN}} & \text { OGM-GE} \citep{peng2022balanced}& \text {ResNet-18} & - & $62.20^*$ \\
        & & \text {MSLR} \cite{yao2022modality} & \text{ResNet-18} & - & $64.42^*$\\ 
        & & \text {PMR} \cite{fan2023pmr} & \text{ResNet-18} & - & $65.30^*$\\
        & & \text {AGM} \cite{li2023boosting} & \text{ResNet-18} & - & $70.16^*$\\
        \cmidrule{2-6}
        & \multirow{4}{*}{\text{Multi-modal SNN}} & \text { WeightAttention}~\cite{liu2022event} & \text {Spiking Transformer} & 4 & $ 64.78$ \\ 
        & & \text{SCA}~\cite{guo2023transformer} & \text{Spiking Transformer} & 4 & $66.53$\\
        & & \text{CMCI}~\cite{jiang2023cmci} & \text{Spiking Transformer} & 4 & $70.02$\\
        & & \cellcolor{gray!10} \text { S-CMRL}  \text {(Ours) } & \cellcolor{gray!10}\text {Spiking Transformer} & \cellcolor{gray!10}4 & \cellcolor{gray!10}$\mathbf{73.25}$ \\
        \midrule
        \multirow{4}{*}{\text { UrbanSound8K-AV }} & \multirow{4}{*}{\text{Multi-modal SNN}} & \text { WeightAttention}~\cite{liu2022event} & \text {Spiking Transformer} & 4 & $ 93.11^*/97.60$ \\ 
        & & \text{SCA}~\cite{guo2023transformer} & \text{Spiking Transformer} & 4 & $96.85^*/97.44$\\
        & & \text{CMCI}~\cite{jiang2023cmci} & \text{Spiking Transformer} & 4 & 97.90\\
        & & \cellcolor{gray!10} \text { S-CMRL}  \text {(Ours) } & \cellcolor{gray!10}\text {Spiking Transformer} & \cellcolor{gray!10}4 & \cellcolor{gray!10}$ \mathbf{98.13}$ \\
        \midrule
        \multirow{4}{*}{\text { MNISTDVS-NTIDIGITS }} & \multirow{4}{*}{\text{Multi-modal SNN}} & \text { WeightAttention}~\cite{liu2022event} & \text {Spiking Transformer} & 16 & $ 99.14$ \\ 
        & & \text{SCA}~\cite{guo2023transformer} & \text{Spiking Transformer} & 16 & 98.98\\
        & & \text{CMCI}~\cite{jiang2023cmci} & \text{Spiking Transformer} & 16 & 99.04\\
        & & \cellcolor{gray!10} \text { S-CMRL}  \text {(Ours) } & \cellcolor{gray!10}\text {Spiking Transformer} & \cellcolor{gray!10}16 & \cellcolor{gray!10}$ \mathbf{99.28}$ \\
        \bottomrule
    \end{tabular}
    }
  \end{threeparttable}
  \caption{Comparison of S-CMRL with state-of-the-art methods on three datasets. The symbol (*) denotes results reported in reference papers, while others are reproduced for fair evaluation.}
  \label{sota}
\end{table*}

\subsection{Experimental Settings}


For visual preprocessing, we employ a direct coding method for still images, where static images are replicated multiple times according to the time steps of the SNN to maintain temporal consistency. Images in the CREMA-D and UrbanSound8K-AV datasets are resized to 128 $\times$ 128 pixels, with random cropping applied for data augmentation. For event-based visual data, we aggregate all event streams into frame representations and resize them, where images in the MNISTDVS dataset are resized to 26 $\times$ 26 pixels to ensure compatibility with the model architecture. 

For audio preprocessing, all raw waveforms are first normalized and resampled to 22,050 Hz to standardize input dimensions across datasets. The preprocessed signals are then transformed into time-frequency representations using the Short-Time Fourier Transform (STFT) with an FFT window length of 512 and a hop length of 353 to generate 2D spectrograms. The resulting amplitude spectrograms undergo logarithmic transformation with an offset of 1e-7 to enhance feature representation and are resized to match the corresponding image dimensions to ensure uniform multimodal input.

All experiments are based on Brain-Cog~\cite{Zeng2023} framework and are conducted on a single NVIDIA A100 GPU. The model is optimized using the Adam optimizer~\cite{Adam2015} with an initial learning rate of $5 \times 10^{-3}$. The training process spans 100 epochs, with a batch size of 128. For the LIF neuron configuration, the initial membrane potential is set to 0, the firing threshold is fixed at 1, and the simulation time step is set to 4. To enable effective gradient backpropagation through spiking neurons, We adapt the Sigmoid function $\operatorname{Sigmoid}(x)=1/(1+\exp (-\alpha x))$ with the parameter $\alpha=4$ as the neuron's surrogate gradient. 

\subsection{Comparison with the State-of-the-Art}


We evaluate the proposed semantic-alignment cross-modal residual learning framework using the Spiking Transformer network on the CRMEA-D and UrbanSound8K-AV datasets, and compare it with existing cross-modal fusion methods such as WeightAttention~\cite{ liu2022event}, SCA~\cite{guo2023transformer} and CMCI~\cite{jiang2023cmci}. Since none of the existing methods are publicly available in code implementation, we reproduce them in the same experimental configuration and mark asterisks in the results from the original papers for comparison. The experimental results are shown in Table~\ref{sota}, demonstrating that our proposed method achieves state-of-the-art performance across all datasets.

Specifically, the S-CMRL method achieves the 73.25\% accuracy on the CREMA-D dataset, compared to other multimodal SNN methods such as WeightAttention (64.78\%), SCA (66.53\%), and CMCI (70.02\%), which have lower performance than S-CMRL. This significant performance improvement validates the effectiveness of the proposed method. Notably, the 73.25\% accuracy achieved by the S-CMRL method also outperforms the artificial neural network based method, which demonstrates the efficiency of the Spiking Transformer network model in integrating audiovisual information.

On the UrbanSound8K-AV dataset, the S-CMRL method achieves an accuracy of 98.13\%, outperforming other multimodal SNN methods, including CMCI (97.90\%), WeightAttention (97.60\%), and SCA (97.44\%). This result indicates that S-CMRL exhibits stronger generalization ability, effectively capturing robust cross-modal representations across different environmental soundscapes.

Furthermore, we evaluated the S-CMRL method on the neuromorphic MNISTDVS-NTIDIGITS dataset. The results show that S-CMRL achieved an accuracy of 99.28\%, surpassing WeightAttention (99.14\%), SCA (98.98\%), and CMCI (99.04\%), which further demonstrates the method’s effectiveness across a broader range of audiovisual datasets.

\begin{figure}
	\centering
	\subcaptionbox{Image with Gaussian noise. \label{fig: resulta}}
	{\includegraphics[width=0.49\linewidth]{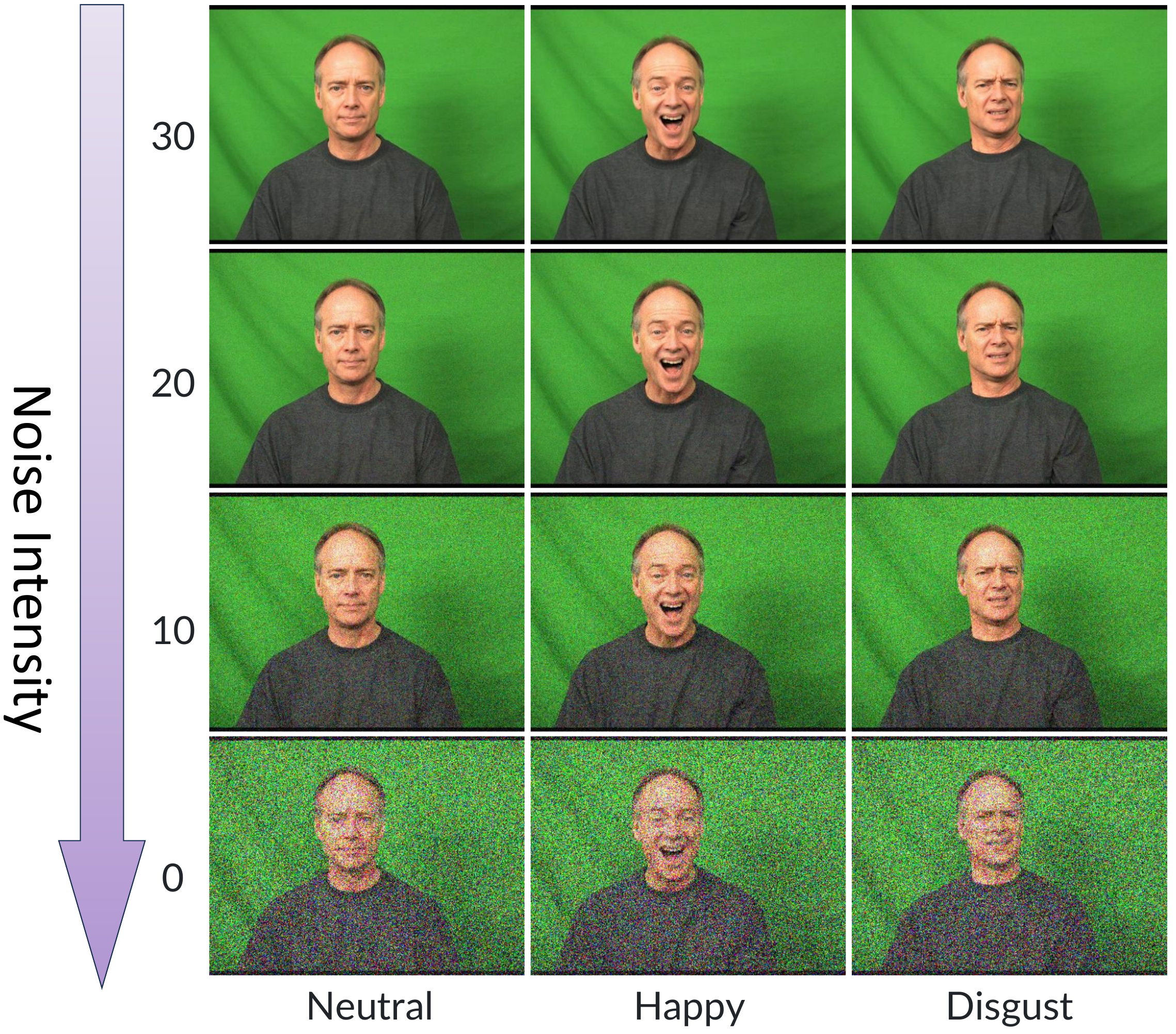}}
	\subcaptionbox{Audio with Gaussian noise. \label{fig: resultb}}
	{\includegraphics[width=0.49\linewidth]{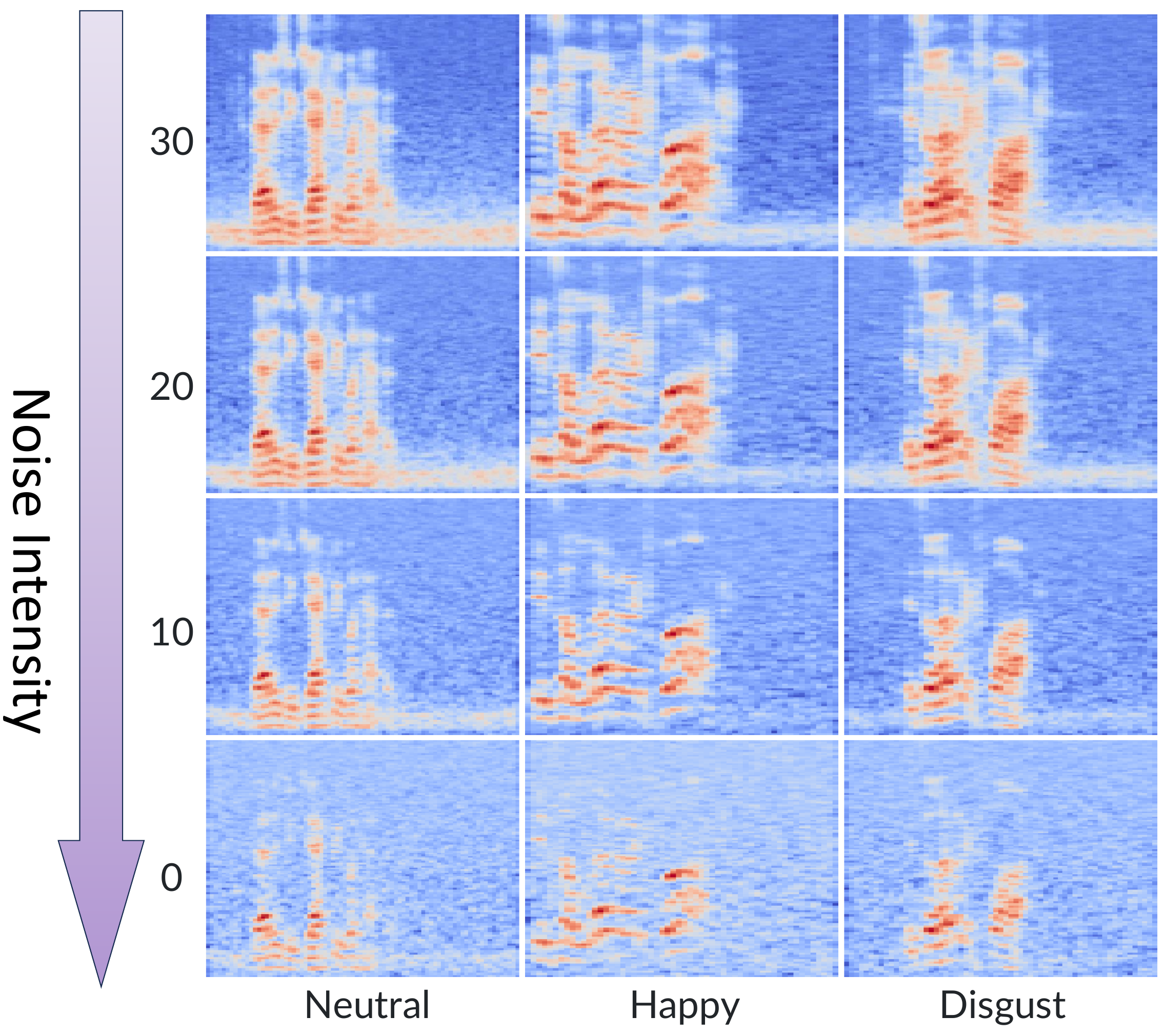}}
	\caption{ 
  Visualization of visual and audio data under different noise intensity in CRMEA-D dataset. The vertical coordinate of each sub-figure represents the SNR value, the smaller the value, the higher the noise intensity. From top to bottom, the graphs present the variation of noise intensity from low to high (30 to 0). The horizontal coordinates show three different emotion categories: Neutral, Happy and Disgust.
  }
  \label{fignoise}
\end{figure}

\subsection{Noise Robustness}
To assess the robustness of S-CMRL to noise, we evaluate the proposed model in terms of noise resistance and compare it with existing multimodal fusion methods. Specifically, we add noisy Gaussian white noise $n$ to the original visual image $\boldsymbol{x}^v$ and audio $\boldsymbol{x}^a$ to obtain the signal input with noise. The added Gaussian white noise follows:
\begin{equation}
  n = \sqrt{\frac{\mathbb{E}[x^2]}{10^{\text{SNR}/10}}} \cdot \mathcal{N}(0,1),
\end{equation}
where $\mathbb{E}[x^2]$ is the mean square value (i.e., signal power) of the original modal input $x$, $\text{SNR}$ is the signal-to-noise ratio in decibels (dB), and $\mathcal{N}(0, 1)$ denotes the normal distribution with a mean of zero and variance of one.

We visualize the added noise to show more intuitively the effect of noise on the image and audio data. Fig.~\ref{fignoise} presents the visual and audio data under different noise intensities. Specifically, Fig.~\ref{fignoise}(a) presents the visual data and Fig.~\ref{fignoise}(b) presents the audio data. As can be seen from the figure, when the signal-to-noise ratio (SNR) is less than or equal to 20, the noise interferes more significantly with the original data.

\begin{figure}[t]
	\centering
		\includegraphics[width=0.95\linewidth]{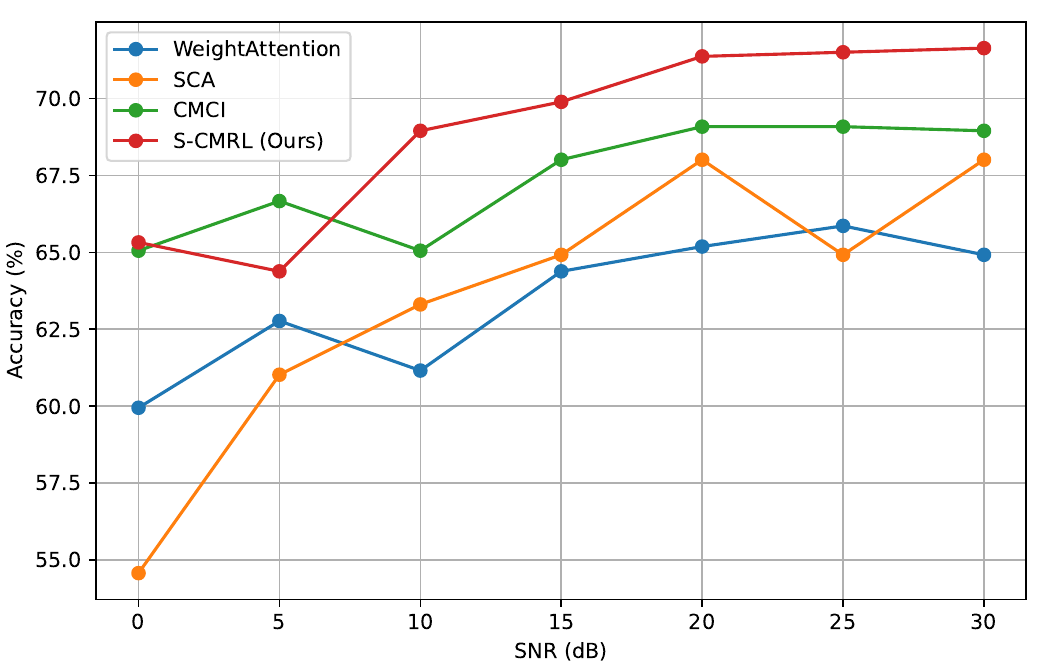}
	\caption{Model accuracy of different multimodal fusion methods for different noise intensities. }
	\label{snr}
\end{figure}

We compare S-CMRL with WeightAttention, SCA, and CMCI under different noise intensities, with results depicted in Fig.~\ref{snr}. The results show that the proposed method obtains the best performance under most noise conditions, exhibits high robustness, effectively reduces the influence of noise on signal processing, and verifies the effectiveness.

\subsection{Ablation study}
In this section, we conduct experiments to verify the validity of each part of the proposed method.

\subsubsection{\bf 
The effectiveness of cross-modal complementary spatio-temporal spiking attention mechanisms
} 
To validate the effectiveness of obtaining complementary features through spatio-temporal attention and using cross-modal complementary features as residual fusion, we compare the cross-modal complementary spatio-temporal attention mechanism with three variants: 1) without using any complementary features, i.e., the visual and audio features are fed directly into the network for audio-visual integration 2) using the cross-modal complementary spatial attention mechanism, obtaining cross-modal spatial complementary features for use as residual fusion 3) Using cross-modal complementary temporal attention mechanism to obtain cross-modal temporal complementary features for residual fusion.

\begin{table}[t]
  \centering
  \begin{threeparttable}
    \resizebox{0.95\linewidth}{!}{
    \begin{tabular}{ccc}
        \toprule
        Datasets & Methods & Accuracy\\
        \midrule
        \multirow{4}{*}{CRMEA-D} & w/o CCSSA & 69.62\%\\
        & w/ CCSSA-Spatial-only & 70.70\%\\
        & w/ CCSSA-Temporal-only & 70.83\%\\
        & w/ CCSSA-Spatiotemporal & \textbf{72.72}\%\\
        \midrule
        \multirow{4}{*}{UrbanSound8K-AV} & w/o CCSSA & 97.44\%\\
        & w/ CCSSA-Spatial-only & 97.82\%\\
        & w/ CCSSA-Temporal-only & 97.79\%\\
        & w/ CCSSA-Spatiotemporal & \textbf{98.05}\%\\
        \bottomrule
    \end{tabular}}
  \end{threeparttable}
  \caption{Ablation experiments with cross-modal complementary spatio-temporal attention mechanisms.}
  \label{Ablation_Variation}
\end{table}

The experimental results are shown in Table \ref{Ablation_Variation}, and it can be seen that the performance of the network decreases dramatically without the use of the cross-modal complementary spatial-temporal attention mechanism, especially in the REMA-D dataset, where the accuracy decreases by about 3\%, which suggests that the CCSSA plays a key role in integrating the complementary information in the feature fusion process. Further, using either the spatial attention mechanism or the temporal attention mechanism alone improves the performance compared to not using complementary features, but relying on only one of the attention mechanisms provides limited improvement. This suggests that while complementary information in the spatial and temporal dimensions each has a contribution to make, it is difficult to fully capture the complex relationships in multimodal data with complementary information in a single dimension. Most importantly, the cross-modal complementary spatio-temporal attention mechanism that combines both spatial and temporal dimensions is able to achieve the greatest performance enhancement, which verifies that CCSSA is able to more comprehensively integrate the complementary information between different modalities and significantly enhance the richness and expressiveness of the feature representation.

\begin{table}[t]
  \centering
  \begin{threeparttable}
    \resizebox{0.9\linewidth}{!}{
    \begin{tabular}{ccc}
        \toprule
        Datasets & Methods & Accuracy\\
        \midrule
        \multirow{2}{*}{CRMEA-D} & w/ CCSSA w/o SAO & 72.72\%\\
        & w/ CCSSA w/ SAO & \textbf{73.25}\%\\
        \midrule
        \multirow{2}{*}{UrbanSound8K-AV} & w/ CCSSA w/o SAO & 98.05\%\\
        & w/ CCSSA w/ SAO & \textbf{98.13}\%\\
        \bottomrule
    \end{tabular}}
  \end{threeparttable}
  \caption{Ablation experiments for semantic alignment optimization mechanisms.}
  \label{Ablation_Semantic}
\end{table}

\subsubsection{\bf 
Effectiveness of Semantic Alignment Optimization Mechanisms}
In order to verify the effectiveness of the semantic alignment optimization mechanism, we added the SAO mechanism based on the introduction of the cross-modal complementary spatio-temporal attention mechanism (CCSSA), respectively, and conducted experimental comparisons on two datasets (CRMEA-D and UrbanSound8K-AV). The experimental results are shown in Table~\ref{Ablation_Semantic}. From Table~\ref{Ablation_Semantic}, it can be seen that in the CRMEA-D dataset, the addition of the semantic alignment optimization mechanism improves the accuracy of the model from 72.72\% to 73.25\%; in the UrbanSound8K-AV dataset, the introduction of the semantic alignment optimization mechanism also brings about a performance enhancement to the model. This indicates that the SAO mechanism plays a key role in enhancing the semantic consistency and complementarity of cross-modal features, and verifies that the SAO mechanism can effectively optimize the feature alignment in the semantic space, thus improving the overall performance of multimodal fusion.

\begin{figure}
	\centering
	\subcaptionbox{CRMEA-D dataset. \label{fig: resulta}}
	{\includegraphics[width=0.49\linewidth]{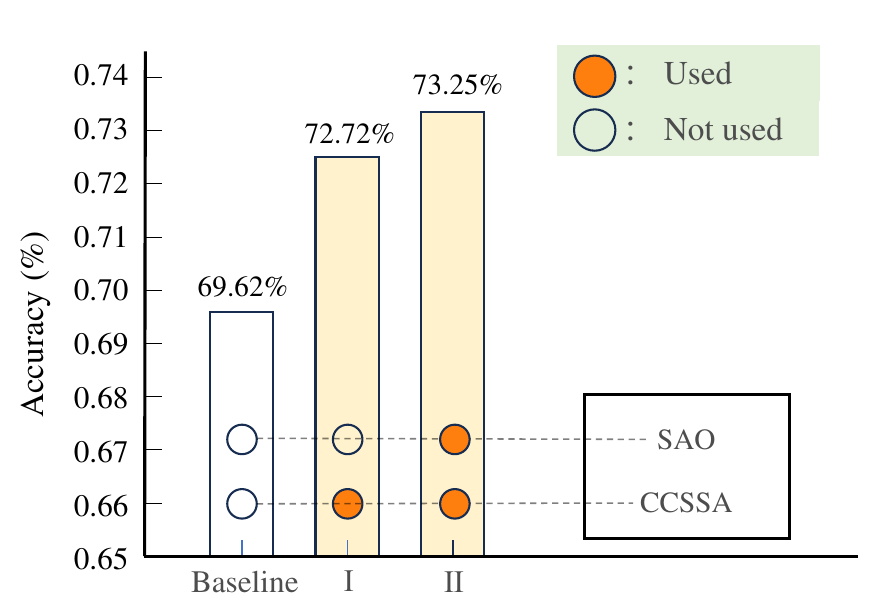}}
	\subcaptionbox{UrbanSound8K-AV dataset.\label{fig: resultb}}
	{\includegraphics[width=0.49\linewidth]{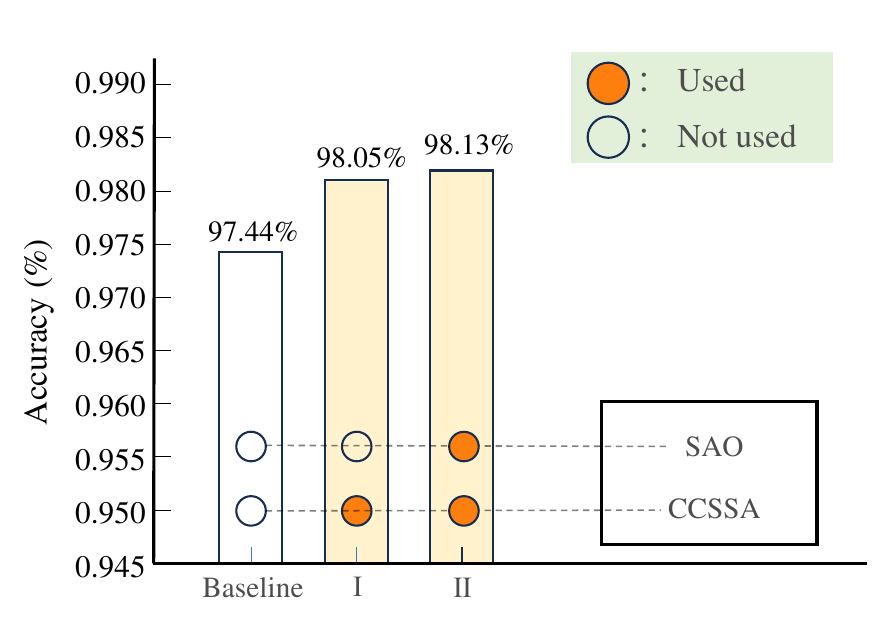}}
	\caption{ 
  Overview of the ablation experiment results for two datasets: CREMA-D (a) and UrbanSound8K-AV (b). Each bar chart shows the accuracy under different experimental settings.
  }
  \label{fig_ablation}
\end{figure}

\subsubsection{\bf 
Summary of ablation experiment results
}
In order to clearly demonstrate the contribution of each part of our proposed method to the model performance, we summarize the experimental results of the two methods in Fig.~\ref{fig_ablation}. It can be seen that using CCSSA alone improves the model performance with an accuracy of 72.72\% on the CREMA-D dataset. Compared with the baseline results (i.e., 69.62\% accuracy achieved without using any of our methods), the network accuracy is improved by about 3\%, validating the effectiveness of our methods. Furthermore, by adding semantic alignment optimization to CCSSA, our method achieves the best result, i.e., 73.25\%. Experimental results on the UrbanSound8K-AV dataset demonstrate a similar improvement, further proving the superiority of our method.

\begin{figure}[t]
	\centering
		\includegraphics[width=0.9\linewidth]{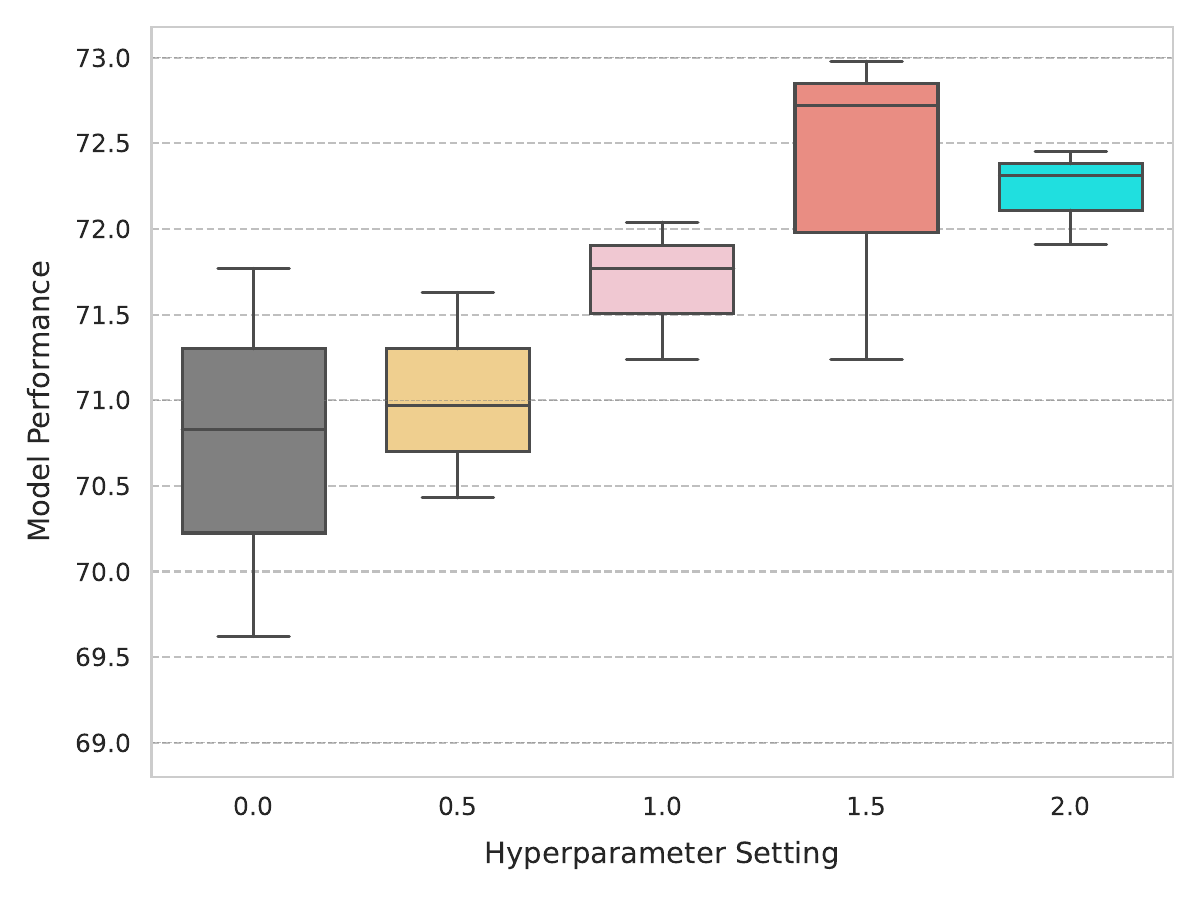}
	\caption{Experimental results of hyperparameter settings on CRMEA-D dataset. The middle line of each box indicates the median. The model performs best when $\alpha$ is set to 1.5.}
	\label{fig_hyperparameter}
\end{figure}

\section{Discussion and Analysis}
\subsection{Hyperparameter Settings}


We use cross-modal complementary spatio-temporal spiking attention to obtain cross-modal complementary features and fuse them with the original features as residuals, and control the fusion strength of the complementary information through the hyperparameter $\alpha$, as shown in Eqs.~\ref{eq14} and \ref{eq15}. In order to demonstrate the effect of cross-modal complementary feature fusion strength on the model performance under different $\alpha$ values, we compare the experimental results under different $\alpha$ values. Fig.~\ref{fig_hyperparameter} displays the average model accuracy across three different random seeds.
When $\alpha=0.0$, only unimodal information is used, and no cross-modal information is interacted. It can be seen that the accuracy of the model is improved whenever cross-modal complementary features are added, which indicates that the cross-modal complementary features enrich the feature information of the original modality. In the CRMEA-D dataset, the optimal performance occurs at a moderate $\alpha=1.5$. When the fusion intensity is further increased, i.e., $\alpha=2.0$, the model performance is relatively degraded, which indicates that the cross-modal complementary features cannot fully reflect the input feature information, and further validates the necessity of retaining the original features.

\subsection{Unimodal Gain Analysis}


In our approach, we propose a cross-modal complementary spatiotemporal spiking attention mechanism. This mechanism acquires complementary information between different modalities across both spatial and temporal dimensions, then integrates this information as “residuals” into the dedicated features of each original modality, thereby significantly improving overall model performance. To verify how cross-modal features enhance single-modality performance, we first train a multimodal model, then extract the model weights for each unimodal branch and use it as the initial weights for that modality in an independent unimodal training.

The experimental results are shown in Fig~\ref{multimodal_finetune}. It shows that the accuracy of the unimodal model is improved after incorporating the cross-modal features, which indicates that the unimodal can benefit from the complementary features of the cross-modal modality, and verifies the effectiveness of the proposed fusion strategy. However, in the CRMEA-D dataset, the accuracy of the audio modality is slightly decreased (from 65.86\% to 64.11\%) after fine-tuning with the weights obtained from the multimodal training, attribute to the large discrepancy between the audio and visual modalities in the CRMEA-D dataset (compared to the 65.65\% accuracy of the audio modality, the accuracy of the visual modality is 43.15\%). This result re-emphasizes the importance of preserving unimodal unique features during cross-modal fusion to avoid performance degradation due to inter-modal differences.

\begin{table}[!t]
  \centering
  \begin{threeparttable}
    \resizebox{1.0\linewidth}{!}{
      \begin{tabular}{ccccc}
        \toprule
        \multirow{2}{*}{Methods} & \multicolumn{2}{c}{CRMEA-D} &\multicolumn{2}{c}{UrbanSound8K-AV}  \\
        \cmidrule(lr){2-3} \cmidrule(lr){4-5}
        & audio & visual & audio & visual \\
        \midrule
        w/o S-CMRL & $\mathbf{65.86}$ & $43.15$ & $91.11$ & $87.63$ \\
        w/ S-CMRL & $64.11_{-1.75}$ & $\mathbf{46.24}_{+3.09}$ & $\mathbf{92.67}_{+1.56}$ & $\mathbf{88.97}_{+1.34}$\\
        \bottomrule
    \end{tabular}
    }
  \end{threeparttable}
  \caption{  
    Information gain from multimodality to unimodality. 
    The subscript indicates the improved accuracy with respect to the baseline shown in the 1st row.}
  \label{multimodal_finetune}
\end{table}

\subsection{Qualitative Visualization Analysis}
To evaluate the effectiveness of our method in learning cross-modal complementary features and providing effective feature information for unimodal tasks, we use the Grad-CAM++~\cite{chattopadhay2018grad} visualization method. This method is able to highlight the local regions of the original image that contribute most to the model's final classification decision. Ideally, incorporating audio features should cause the visual part to focus more on the sound-producing regions. For example, in the ``dog barking'' category, the visual attention should be concentrated on the dog's mouth, which helps the model better understand the source and related features of the sound.

\begin{figure}[t]
	\centering
		\includegraphics[width=1.0\linewidth]{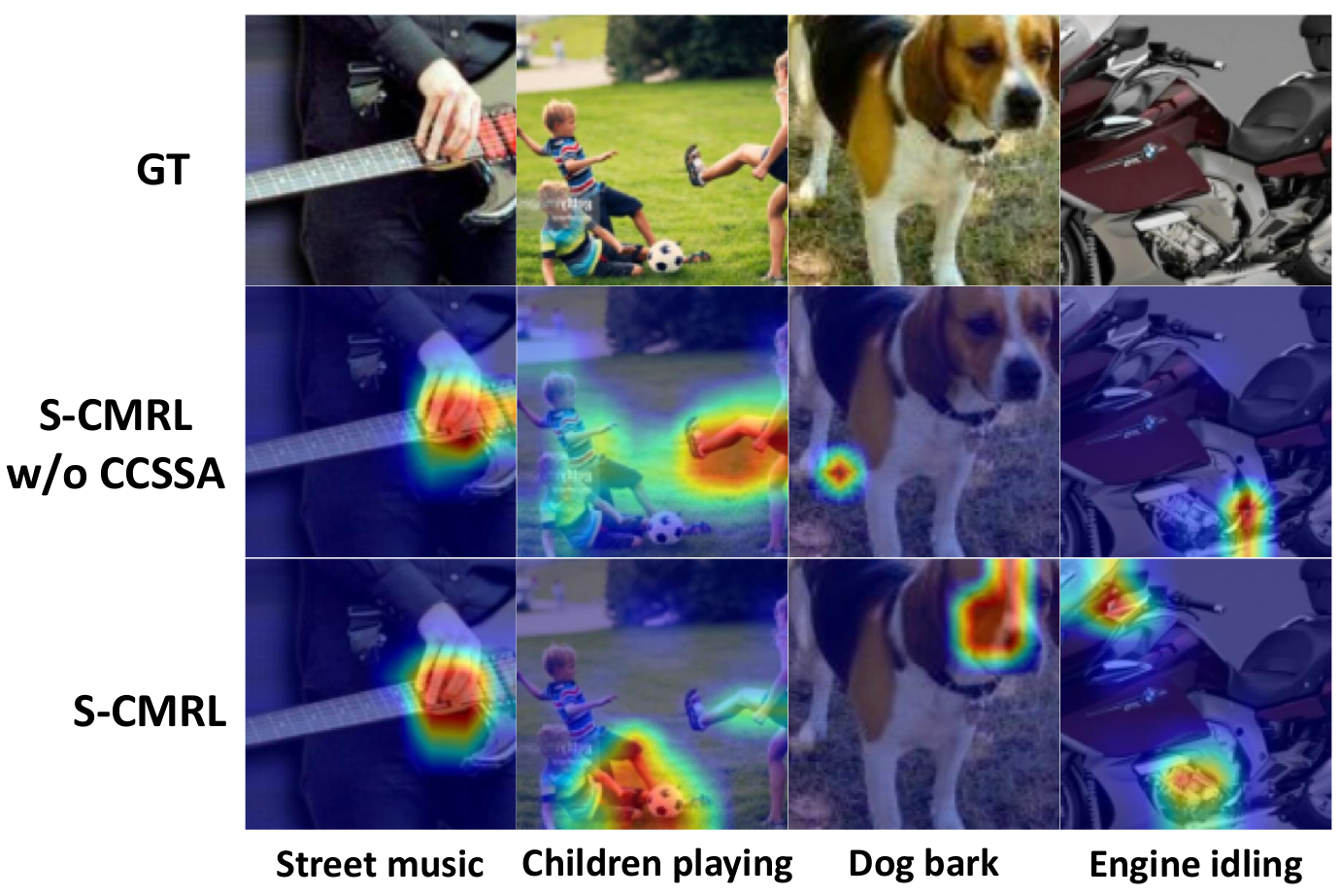}
	\caption{
  Class activation mapping in the UrbanSound8K-AV dataset, with four categories selected for presentation. The top row shows the ground truth of the original image, the middle row shows the effect of our method (S-CMRL) without CCSSA, and the bottom row shows the effect of S-CMRL.}
	\label{gradcam}
\end{figure}

To comprehensively evaluate the effectiveness of our method, we compare three different settings, as shown in Fig.~\ref{gradcam}: The first row displays the ground truth of the original images, serving as the baseline for comparison. The second row shows the effect of our method, S-CMRL without CCSSA. In this row, we observe that although the model still pays attention to some regions, compared to the ideal case, the visual part lacks a focused attention on key information. Specifically, in the ``Dog bark'' category, the attention is more dispersed, failing to focus on the dog's mouth. This indicates that in the absence of CCSSA, the model fails to effectively leverage audio information to guide visual attention, thereby affecting the extraction of key features and classification decisions.

In contrast, the bottom row shows the effect of the S-CMRL method. In this case, the model effectively integrates audio and visual information, significantly enhancing the attention to key features, such as the dog's mouth and the engine in the motorcycle. This result demonstrates that the incorporation of audio features and the learning of cross-modal complementary features allow the model to focus more on regions relevant to sound in the visual image, thus significantly improving the accuracy of classification. This validates the effectiveness of our method in cross-modal feature fusion, especially in terms of its ability to focus on key sound-related areas.

\section{CONCLUSION}
This paper proposes a semantic-alignment cross-modal residual learning (S-CMRL) framework, a Transformer-based multimodal spiking neural network that effectively integrates visual and auditory modalities. By introducing a cross-modal complementary spatiotemporal spiking attention mechanism, S-CMRL extracts complementary features across both spatial and temporal dimensions and incorporates them as residual connections into the original features, thereby enhancing the richness and expressive capacity of feature representations. Furthermore, a semantic alignment optimization mechanism aligns cross-modal features in the semantic space, further improving their consistency and complementarity. Experimental results indicate that S-CMRL surpasses existing methods on multiple public datasets including CREMA-D, UrbanSound8K-AV, and MNISTDVS-NTIDIGITS, and demonstrates strong robustness under noisy conditions. Ablation studies validate the critical role of the proposed mechanism in boosting model performance. This work provides a novel and effective approach for feature fusion and representation in multimodal spiking neural networks, underscoring the potential of SNNs in handling complex multimodal tasks.

\bibliographystyle{IEEEtran}
\bibliography{New_IEEEtran_how-to}

\end{document}